\pdfoutput=1

\documentclass[11pt]{article}

\usepackage[preprint]{acl}
\usepackage{xcolor}
\usepackage{hyperref}
\usepackage{times}
\usepackage{latexsym}
\usepackage{soul}
\usepackage{amsmath}
\usepackage{booktabs}
\usepackage{multirow}
\usepackage{microtype}
\usepackage{url}
\usepackage{lineno}
\usepackage{graphicx} 
\usepackage{enumitem}
\usepackage{titlesec}
\usepackage{rotating}
\usepackage{amssymb}
\usepackage[T1]{fontenc}

\usepackage[utf8]{inputenc}

\usepackage{microtype}
\definecolor{aoenglish}{rgb}{0.0, 0.5, 0.0}
\usepackage{inconsolata}

\usepackage{graphicx}

\title{Unraveling Interwoven Roles of Large Language Models in Authorship Privacy: Obfuscation, Mimicking, and Verification}

\author{
  Tuc Nguyen$^{1}$, Yifan Hu$^{2}$, Thai Le$^{1}$ \\
  $^{1}$Indiana University \quad
  $^{2}$Northeastern University \\
  \texttt{\{tucnguye, tle\}@iu.edu} \quad \texttt{yif.hu@northeastern.edu}
}

\expandafter\def\expandafter\normalsize\expandafter{%
    \normalsize%
    \setlength\abovedisplayskip{3pt}%
    \setlength\belowdisplayskip{2pt}%
}

\begin{document}
\maketitle
\begin{abstract}
Recent advancements in large language models (LLMs) have been fueled by large-scale training corpora drawn from diverse sources such as websites, news articles, and books. These datasets often contain \textit{explicit} user information, such as person names, addresses, that LLMs may unintentionally reproduce in their generated outputs. Beyond such explicit content, LLMs can also leak identity-revealing cues through \textit{implicit} signals such as distinctive writing styles, raising significant concerns about authorship privacy. There are three major automated tasks in authorship privacy, namely authorship obfuscation (AO), authorship mimicking (AM), and authorship verification (AV). Prior research has studied AO, AM, and AV independently. However, their interplays remain under-explored, which leaves a major research gap, especially in the era of LLMs, where they are profoundly shaping how we curate and share user-generated content, and the distinction between machine‑generated and human‑authored text is also increasingly blurred. This work then presents the first unified framework for analyzing the dynamic relationships among LLM-enabled AO, AM, and AV in the context of authorship privacy. We quantify how they interact with each other to transform human‑authored text, examining effects at a single point in time and iteratively over time. We also examine the role of demographic metadata, such as gender, academic background, in modulating their performances, inter-task dynamics, and privacy risks. All source code will be publicly available.

\end{abstract}

\section{Introduction}
\begin{figure}[tb]
    \centering
    \includegraphics[width=0.475\textwidth]{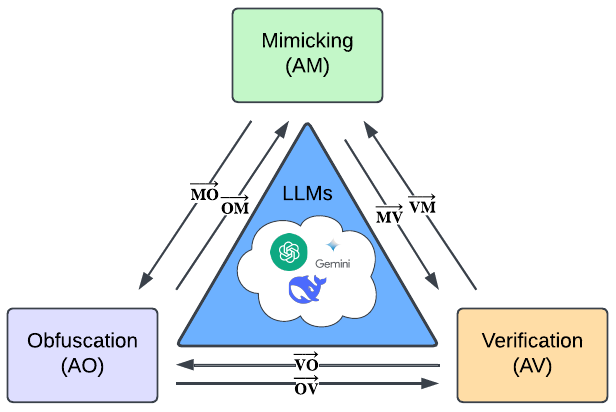}
    \caption{The interactive influence loop between LLMs, obfuscation, mimicking, and verification.}
    \label{motivation}
    \vspace{-20pt}
\end{figure}

Recent advances in LLMs have been extraordinary, driven largely by the massive amounts of training data indiscriminately sourced from diverse online platforms such as websites, news outlets, and books \cite{brown2020language, le2023bloom, touvron2023llama, achiam2023gpt}.
This training data often includes extensive writing contributions by the same authors, publicly shared across various platforms \cite{gao2020pile, raffel2020exploring}. These sources frequently contain \textit{explicit} user information, such as names, addresses, and phone numbers, which LLMs can inadvertently expose during their text generation process \cite{weidinger2021ethical, kim2024propile}.
Beyond explicit details, user identification can also be inferred from \textit{implicit} information, such as their distinctive writing styles, that does not immediately give out the authors' identities. 
Research in human cognitive science and linguistics highlights that individual backgrounds significantly shape writing styles \cite{zheng2006framework, cheng2023marked, deshpande2023toxicity, xing2024alison, he2024cos}, facilitating bidirectional inferences between implicit information (e.g., writing style) and explicit information (e.g., names, ages, or areas of expertise).
Recent studies also reveal that text generated by LLMs can also capture human personality traits \cite{karra2022estimating, jiang2024evaluating, jiang2023personallm, bang2024measuring, an2024measuring}, and vice versa--i.e., explicit information about specific individuals or groups can be used by LLMs to produce personalized outputs or \textit{mimic individuals' writing styles}~\cite{chen2024large, salemi2023lamp}.

Although the authorship mimicking (AM) capabilities of LLMs--i.e., their ability to replicate an individual's writing style, are impressive, this capability could also enable malicious activities, such as impersonating public figures to spread misinformation or commit fraud~\cite{deshpande2023toxicity, jiang2024evaluating}. 
For instance, a fraudster could fine-tune an LLM on publicly available texts authored or spoken by a target victim (e.g., social media posts, interviews) and prompt LLMs to generate spam emails or persuasive messages that pretend to be delivered by the victim \cite{salewski2023context}. Contrasting with AM, authorship obfuscation (AO)~\cite{uchendu2024catch} aims to conceal an author’s identity by altering stylistic features of text while preserving its original meaning. By masking writing style before public dissemination (e.g., on social media), AO can help protect whistleblowers, such as writers or speakers, from potential anonymity exposure. 
In addition, authorship verification (AV) is the process of determining the author of a particular piece of writing. AV poses significant privacy risks by enabling the de-anonymization of individuals through their writing style, which can facilitate surveillance, behavioral profiling, and misuse without informed consent. 

While AO, AM, and AV have each been studied in isolation, their interactions within a unified framework remain underexplored or limited to only specific pairwise formulations, such as AV and AO in the context of LLM-generated text ~\citet{uchendu2023attribution}. 
In addition, real-world scenarios often involve multiple rounds of text transformation, where content is repeatedly mimicked, obfuscated, and verified—either by different LLMs or within multi-turn dialogue settings where LLMs interact with one another \cite{duan2023botchat}.
To address this gap, our study investigates three key scenarios in which LLMs play triple roles in authorship privacy (Fig.~\ref{motivation}), analyzing their individual effects, interdependencies, and collaborative influences. Understanding the interplay among these capabilities is crucial for netizens in today's LLM era, where users may rely on LLMs to obfuscate their writing style, while others may utilize LLMs to recover or attribute the original authorship. \textbf{Our contributions} include: (1) the first unified framework for studying the bidirectional effects among AO, AM, and AV; (2) empirical findings revealing distinct task-specific strengths of various commercial LLMs; (3) detailed analysis showing how demographic and metadata influence these interactions. Our analysis shows that obfuscation tends to outperform mimicking in interactive settings, effectively disrupting authorial signals. However, mimicking can partially reverse obfuscation over successive cycles, gradually restoring aspects of the original writing style. Furthermore, models with stronger reasoning abilities (e.g., o3-mini, Deepseek) according to the benchmark \footnote{https://www.vals.ai/benchmarks/math500-05-09-2025}, excel at verification and concealing authorial traits but are less effective at faithfully replicating an author's distinctive style.
 
\section{Related Works}
Beyond explicit metadata leakage such as names, social security numbers, LLMs' generations can also reflect \textit{implicit and private authorship signals} such as writing style, tone, or rhetorical structure, many of which are uniquely identifiable to specific individuals \cite{zheng2006framework, cheng2023marked, deshpande2023toxicity, xing2024alison, he2024cos}. Thus, these models may memorize and reproduce identifiable features of authorship through their generated texts, so-called {AM}, introducing interesting interwoven relationships with LLM-enabled {AO} and {AV}. 

\vspace{3pt}
\noindent \textbf{Authorship Obfuscation (AO)} hides the original author's identity by altering stylistic cues without compromising semantic content. Recent methods include ALISON \cite{xing2024alison}, which performs obfuscation by substituting stylistic sequences, and StyleRemix~\cite{fisher2024styleremix}, which utilizes AdapterMixup~\cite{nguyen2024adapters} to train adapters for various stylistic dimensions and mix them. Different prompting-based approaches using LLMs have also been proposed \cite{hung2023wrote, pape2024prompt}.

\vspace{3pt}
\noindent \textbf{Authorship Mimicking (AM)} is the reverse of AO, aiming to generate text in the style of a specific author. LLMs excel in this task due to their few-shot and in-context learning capabilities, raising ethical concerns around impersonation, misinformation, and malicious use \cite{deshpande2023toxicity, jiang2024evaluating}. Recent work has shown that LLMs can be fine-tuned or prompted to convincingly replicate individual writing styles from publicly available content \cite{salewski2023context}, making these capabilities intersect with privacy risks, such as when the LLMs leak memorized training examples \cite{carlini2022quantifying, zhang2023counterfactual}.

\vspace{3pt}
\noindent \textbf{Authorship Verification (AV)} seeks to determine or confirm whether a given text was written by a particular author, based on linguistic cues or stylistic fingerprints \cite{huang2025authorship}. With the advancement of model size scaling laws, LLMs can now perform AV in few-shot settings \cite{hung2023wrote, huang2024can}.

\vspace{3pt}
\noindent \textbf{Interdependency of AO, AM, and AV.} Prior research has largely treated AO, AM, and AV in isolation. However, their \textbf{pairwise interactions}, especially under the influence of LLMs, remain underexplored and foundational to many practical scenarios. For instance, for AO–AV, users obfuscate their writing style to protect identity, while adversaries re-identify authorship, creating a privacy-versus-attribution dynamic; for AM–AV, attackers mimic a target author’s style to deceive attribution models, challenging the robustness of verification systems; and for AO–AM, one can attempt to reconstruct authorial style from obfuscated text, testing the boundaries of stylistic recovery. Moreover, AO, AM, and AV can also form a closed loop in a \textbf{triplet-wise interaction}, reflecting how a text authorship changes under the influence of LLMs overtime. Our work is the first to address all pairwise and triplet-wise interdependencies of LLM-enabled AO, AM, and AV.

\section{Research Questions and Formulation}
\subsection{Research Questions}
We propose three \textbf{research questions (RQs) } to investigate both isolated and multi-level interdependencies among LLM-enabled authorship privacy tasks AO, AM, and AV, aiming to understand how individual and joint model behaviors influence the privacy and stylometry--i.e., writing styles, in complex authorship pipelines. Practical implications of our RQs are motivated in Appendix.~\ref{practical_application}.

\begin{enumerate}[label={},leftmargin=\dimexpr\parindent-0.7\labelwidth\relax,noitemsep,topsep=0pt]
    \item \textbf{\textit{\ul{RQ1:}}} How effectively can different LLMs perform AO, AM, AV {\textit{in isolation}}, and which models are best suited for specific goals such as privacy preservation and stylistic imitation?
    \item \textit{\textbf{\ul{RQ2:}}} How do LLM-enabled AO, AM, and AV influence one another to transform individuals' stylometries when used {\textit{in conjunction at one point in time}}, including their pairwise and triplet interactions?
    \item \textbf{\textit{\ul{RQ3:}}} How do LLM-enabled AO, AM, and AV influence one another to transform individuals' stylometries when used {\textit{in conjunction iteratively through time}}?
\end{enumerate}

\noindent To answer these RQs, we first formally define the evaluation of AO, AM, and AV of a target LLM $f(\cdot)$. 
For a given author \( a \), let \( \mathcal{D}_a{=}\{ (x_1, y_1), (x_2, y_2), \dots, (x_n, y_n)\} \) represent a set of \( a \)'s original written documents paired with their corresponding author labels. \( M_a \) denotes the metadata associated with author \( a \), such as name, field of study. We define \( C_a{=}\{ M_a, \mathcal{D}_a\} \) as the context available to $f(\cdot)$. For example, $f^{AO}(x| C_a)$ denotes the output obfuscation text of LLM $f(\cdot)$ on the input text $x$ given the context $C_a$. $d(\cdot)$ is a stylometric distance defined on the two sets of input texts.

\subsection{Isolation - No Interdependency}~\label{sec:isoltion}
We begin by formulating AO, AM, and AV in isolation to evaluate the standalone performance of a specific LLM $f$. This setting is the most common in prior work, where researchers aim to quantify how well an individual LLM performs on specific authorship privacy tasks \cite{hung2023wrote, huang2024can, fisher2024styleremix, pape2024prompt, salewski2023context}.

\vspace{3pt}
\noindent \textbf{AO.} To evaluate the effectiveness of AO on an input text $x$, we compute the distance $d(\cdot)$ between the original authentic texts and the obfuscated one (Eq.~\ref{authorship_obfuscation}). The larger the distance, the more divergent the obfuscated text becomes from the original, suggesting more effective obfuscation.
\begin{align}
    {AO} &= {d}(f^{AO}(x| C_a),\;\mathcal{D}_a)
    \label{authorship_obfuscation}
\end{align}

\vspace{3pt}
\noindent \textbf{AM.} We evaluate the effectiveness of AM on an input text $x$ by computing the distance between the original texts and the mimicked text (Eq.~\ref{authorship_mimicking}). The smaller stylometric distance $d(\cdot)$, the more similar the mimicked text is to the original, suggesting more effective mimicking.
\begin{align}
    {AM} &= {d}(f^{AM}(x| C_a),\;\mathcal{D}_a)
    \label{authorship_mimicking}
\end{align}

\vspace{3pt}
\noindent \textbf{AV.} We evaluate the effectiveness of AV on an input text $x$ by comparing its binary predictive verification --i.e., whether the text was written by author $a$ or by someone else (Eq.~\ref{authorship_verification}). The higher verification accuracy, the more effective $f(\cdot)$ is at correctly identifying the author's text.
\begin{align}
    {AV} &= \mathbb{I}(f^{AV}(x| C_a) == a)
    \label{authorship_verification}
\end{align}

\subsection{Pairwise Interdependency}
Netizens are increasingly relying on LLMs to refine or disguise their writing through polishing, paraphrasing, or rephrasing, before sharing and publishing their content. These scenarios highlight a growing trend in which multiple LLMs are employed within a single pipeline: one model generates or modifies text, while another evaluates or attributes authorship. Consequently, the input to these models is not always original author-written text but may already have undergone AI-driven transformation \cite{uchendu2023attribution}. To better understand these interactions, we conduct \textit{pairwise interdependency} evaluations that measure their bidirectional relationships--i.e., \textit{how one LLM’s capabilities influence the performance of others} (Fig.~\ref{motivation}). To reflect the realistic scenario where the users prefer the best models for specific tasks, we designate a \textit{``judge''} $f_\textit{judge}$ for each task, or the LLM that is selected based on its highest standalone performance in isolation (${\S}$~\ref{sec:isoltion}), for this evaluation.

\vspace{3pt}
\noindent \textbf{Influence of Obfuscation.} We factorize the influence of AO in the authorship pipeline into (1) how AO influences AM ($\overrightarrow{OM}$) and (2) how AO influences AV ($\overrightarrow{OV}$). For $\overrightarrow{OM}$, we first generate the obfuscated versions of an input text $x$, denoted \(x_{\textit{obf}}\), using various LLMs. Each of the obfuscated texts then serve as an input for the mimicking ``judge'' - a ``ground-truth'' LLM with the highest AM performance in isolation (${\S}$~\ref{sec:isoltion}), which attempts to reconstruct the original style of input $x$ (Eq.~\ref{obfuscation_mimicking}). We compare the mimicked outputs to the original, authentic texts. The greater their stylistic divergence is, the more effective the obfuscated input, and hence the more influential the corresponding AO, and vice versa:
\begin{align}
    \overrightarrow{OM} &= {d}(f^{\text{AM}}_{judge}(x| x_\textit{obf}) ,\;\mathcal{D}_a)
    \label{obfuscation_mimicking}
\end{align}
For $\overrightarrow{OV}$, we pass the obfuscated texts \(x_{\textit{obf}}\) to a verification ``judge''. We compute verification accuracy on the original input $x$ given the obfuscated texts (Eq.~\ref{obfuscation_verification}). The lower the accuracy, the more effective the obfuscation is; otherwise, it suggests the author’s style remains identifiable. This evaluation provides a practical measure of AO by testing whether others can still attribute the distorted writing to its original author. Such insights are particularly valuable in privacy‑sensitive settings—e.g., anonymous investigative journalism or whistleblowing—where safeguarding the author's identity is paramount:
\begin{align}
    \overrightarrow{OV} &=  \mathbb{I}(f^{AV}_{judge}(x| x_\textit{obf}) == a)
    \label{obfuscation_verification}
\end{align}

\vspace{3pt}
\noindent \textbf{Influence of Mimicking.} We factorize the influence of AM in the authorship pipeline into (1) how AM influences AO ($\overrightarrow{MO}$) and (2) how AM influences AV ($\overrightarrow{MV}$).
For ($\overrightarrow{MO}$), we first generate mimicking versions of the input text $x$, denoted as $x_{mimic}$, using various LLMs. These mimicked texts then serve as the reference inputs for the obfuscation ``judge''. Then, we compare the resulting obfuscated outputs to the original, authentic texts (Eq.~\ref{mimcking_obfuscation}). Obfuscation style significantly diverging from the originals indicates that the mimicking was effective in replicating the author’s writing style, and vice versa:
\begin{align}
    \overrightarrow{MO} &=  d(f^{AO}_{judge}(x|x_\textit{mimic}), \; \mathcal{D}_a )
    \label{mimcking_obfuscation}
\end{align}

\noindent For $\overrightarrow{MV}$, we feed $x_{\text{mimic}}$ into a verification ``judge''. We calculate the verification accuracy of the predictive author with $x$'s original author $a$ (Eq.~\ref{mimicking_verification}). A high verification accuracy indicates that the mimicked text effectively replicates the original author’s writing style, whereas a low accuracy suggests poor stylistic imitation:
\begin{align}
    \overrightarrow{MV} &=  \mathbb{I}(f^{AV}_{judge}(x| x_\textit{mimic}) == a)
    \label{mimicking_verification}
\end{align}

\vspace{3pt}
\noindent \textbf{Influence of Verification.} We factorize the influence of AV in the authorship pipeline into (1) how AV influences AO ($\overrightarrow{VO}$) and (2) how AV influences AM ($\overrightarrow{VM}$). In other words, AV acts as a filtering process to select only the texts verified as being authored by $a$ as the input \textit{contexts} for AO and AM. Intuitively, AV decides how pure or contaminated $C_a$ is. To do this, we randomly sample $n$ \textit{noisy texts} or documents written by authors different from $a$, supposedly these are imposter samples.
In both settings, we assess AV performance under two conditions: (1) \textit{perfect $C_a$:} where all input context are genuine samples from the target author, and (2) \textit{noisy $\overline{C}_a$:} where we introduce imposter samples from other authors that the model nonetheless classifies as the target author. Persistent positive classification of these imposter texts indicates weaker verification robustness. We then compute the distance of mimicking and obfuscation texts on the original input $x$, with the ground truth samples are all genuine and noisy (Eq.~\ref{verification_o}, Eq.~\ref{verification_m}).
\begin{align}
    \overrightarrow{VO} &= {d}( f^{AO}_{judge}(x|C_a), \;f^{AO}_{judge}(x|\overline{C}_a) )
    \label{verification_o}
\end{align}
\begin{align}
    \overrightarrow{VM} &= {d}( f^{AM}_{judge}(x|C_a), \;f^{AM}_{judge}(x|\overline{C}_a) )
    \label{verification_m}
\end{align}

\subsection{Triplet-wise Interdependency}
While previous evaluations identify which models excel at individual tasks and how they are pairwise-interdependent, this section investigates \textit{the authorship pipeline cycle as a whole} (Fig.~\ref{motivation})--i.e., how AO and AM alter verification accuracy and the linguistic distribution of original human texts. By orchestrating multiple LLMs, each deployed for its strongest capability, whether AO, AM or AV, we evaluate their collective impact on authorship privacy. This integrated perspective mirrors real‑world workflows in which texts undergo successive AI‑mediated transformations, from iterative edits in anonymous online forums to chained paraphrasing and verify in whistleblowing activities.

\section{Experiment Setup}

\renewcommand{\tabcolsep}{2pt}
\begin{table}[tb]
\footnotesize
\centering
\begin{tabular}{lcccccc}
\toprule
\multirow{2}{*}{\textit{\textbf{Models}}} 
& \multicolumn{2}{c}{\textbf{AO}} 
& \multicolumn{2}{c}{\textbf{AM}} 
& \multicolumn{1}{c}{\textbf{AV}} \\
\cmidrule(lr){2-3} \cmidrule(lr){4-5} \cmidrule(lr){6-7}

& {PPL ($\uparrow$)} & {SIM ($\downarrow$)} 
& {PPL ($\downarrow$)} & {SIM ($\uparrow$)} 
& {Acc} ($\uparrow$)\\

\midrule
\textit{4o-mini}         & 0.72 & 0.12 & \textbf{0.65} & \textbf{0.13} & 0.45 &  \\ 
\textit{o3-mini}         & \textbf{2.71} & \textbf{0.10}  & 1.57 & 0.11 & \textbf{0.89} &  \\ 
\textit{deepseek}        & \ul{1.08} & \ul{0.11} & 1.86 & \ul{0.12} & \ul{0.74} &  \\ 
\textit{gemini}          & 0.31 & 0.12 & \ul{1.00} & \textbf{0.13} & 0.39 &  \\ 

\bottomrule
\end{tabular}
\caption{Isolation evaluation on AO, AM, and AV across different models. \textbf{Bold} and \ul{underline} indicate each metric's best and second-best performance, respectively.}
\label{intra_model_evaluation}
\vspace{-10pt}
\end{table}

\noindent \textbf{Models.} We utilize the well-known commercial LLMs of varying presence of reasoning capability and origins: GPT-4o-mini \cite{achiam2023gpt}, GPT-o3-mini \cite{brown2020language}, Gemini-2.0 \cite{team2023gemini}, and Deepseek-v3 \cite{liu2024deepseek}.

\vspace{3pt}
\noindent \textbf{Datasets.} We utilize three datasets: \textit{Speech}: US Presidents' speeches from \citet{fisher2024styleremix}, \textit{Quora}: Quora blog posts by diverse users with active online presence that we collect ourself; and \textit{Essay:} writing essays from layperson~\cite{li2025writes}. These corpora vary in text length and author notoriety, descending from \textit{Speech}, \textit{Quora} and \textit{Essay}. They also allow us to evaluate LLMs' performance on writing by both native and non-native English speakers.

\vspace{3pt}
\noindent \textbf{Prompts.} Following previous works such as LIP~\cite{huang2024can}, we design prompts along four key dimensions: \textit{Context}, \textit{Task}, \textit{Instruction}, and \textit{Output} to characterize open-ended LLMs' behavior systematically~\cite{cheng2023compost}. Specifically, we prompt LLMs to focus on writing style rather than topic or content differences.

\vspace{3pt}
\noindent \textbf{Metrics.} In our work, authorship privacy depends on identifying linguistic traits that are unique to individuals and can also help differentiate human-authored text from that generated by LLMs. Particularly, we examine how \textit{4 key linguistic features} change \textit{before and after} an authorship task AO, AM and AV is performed. Central to this is word distribution, quantified using TF-IDF similarity (denoted as \textbf{SIM}), which is also widely applied in detecting deepfake text by revealing unnatural or overly consistent vocabulary usage \cite{becker2023paraphrase}. Additionally, we evaluate language naturalness using perplexity (denoted as \textbf{PPL}) and also report the \textbf{KL} divergence over the distribution of text PPL scores. This metric is commonly employed to capture the natural writing patterns of individuals and to detect machine-generated text that may appear overly fluent or statistically optimized compared to genuine human writing. In our experimental setup, we conduct evaluations both with metadata \( C_a{=}\{ M_a, \mathcal{D}_a\} \) and without metadata \( C_a{=}\{\mathcal{D}_a\} \).

\noindent\textit{Details of the datasets, prompts, and metrics are provided in the Appendix. \ref{dataset_statistics}, \ref{app_prompt_construction}, \ref{evaluation_metrics}, respectively.}

\section{Experiment Results}
\subsection{Isolation Evaluation (RQ1)}
\label{intra_evaluation}
Overall, \textit{o3-mini} performs the best in AO and AV tasks, and \textit{4o-mini} leads in faithful AM (Table~\ref{intra_model_evaluation}). Particularly, \textit{o3-mini} achieves the highest perplexity (2.71) and lowest similarity (0.10) in AO, indicating more distinct and less traceable outputs. For AM, \textit{4o-mini} excels with the lowest PPL (0.65) and highest similarity (0.13), reflecting better stylistic imitation of the original texts. For AV, \textit{o3-mini} identifies authorships with the highest accuracy (0.89).

\begin{figure}[tb!]
    \centering
    \includegraphics[width=0.42\textwidth]{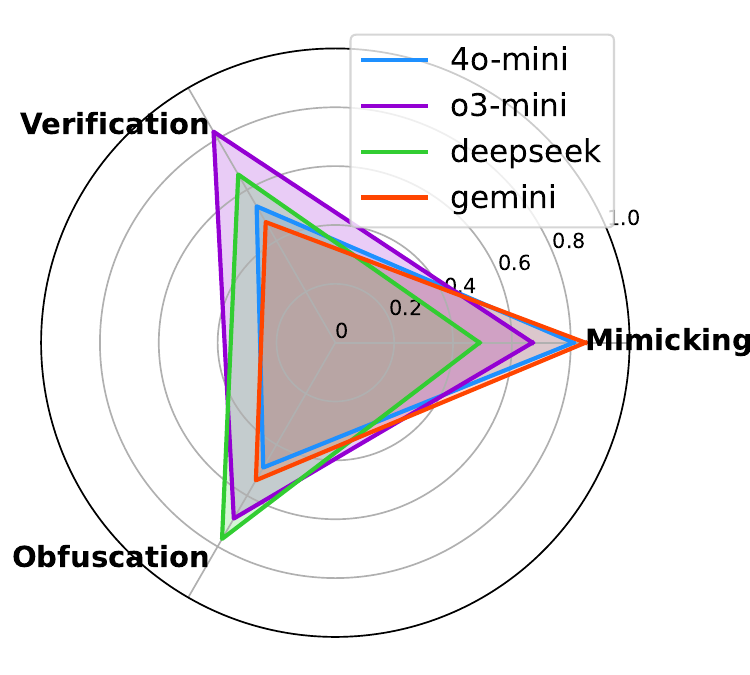}
    \vspace{-10pt}
    \caption{We present an overall pairwise interdependency evaluation of each LLM across the tasks of AO, AM, and AV. For each aspect, the final score is computed as the average across two ``judge'' evaluations to enable relative comparison.}
    \label{overall_intra_evaluation}
    \vspace{-15pt}
\end{figure}

\subsection{Pairwise Interdependency (RQ2)}
\label{pairwise_interdependency}
From the isolation evaluation (Sec.~\ref{intra_evaluation}), we select \textit{o3-mini} as both the obfuscation and verification judge, and \textit{4o-mini} as the mimicking judge to assess the interplays among the authorship tasks.
Fig.~\ref{overall_intra_evaluation} presents a comprehensive comparison of the four models' influence capabilities across AO, AM, and AV. Overall, \textit{gemini}, \textit{deepseek}, and \textit{o3-mini} are the most influential or effective with mimicking, obfuscation, and verification, respectively. We analyze each authorship task in detail as follows.

\renewcommand{\tabcolsep}{1.2pt}
\begin{table}[tb]
\footnotesize
\centering
\begin{tabular}{llcccccccccc}
\toprule
& \multirow{2}{*}{\textbf{Models}} 
& \multicolumn{3}{c}{\textbf{Speech}} 
& \multicolumn{3}{c}{\textbf{Quora}} 
& \multicolumn{3}{c}{\textbf{Essay}} \\
\cmidrule(lr){3-5} \cmidrule(lr){6-8} \cmidrule(lr){9-11}
& & KL & SIM & ACC
  & KL & SIM & ACC
  & KL  & SIM & ACC \\
\midrule
\multirow{4}{*}{\begin{sideways}\textit{w. meta}\end{sideways}} 
& \textit{4o-mini}   &  0.14 & 0.08 & 0.71 & 1.21 & 0.19 & 0.67 & 1.99 & 0.18 & 0.59\\
& \textit{o3-mini}   & \ul{0.91} & \ul{0.08} & \ul{0.58} &\ul{1.96} & \ul{0.16} & \ul{0.59} & \ul{2.15} & \textbf{0.15} & \textbf{0.51}\\
& \textit{gemini}    &  0.23 & 0.09 & 0.66 & 1.82 & 0.18 & 0.69 & 1.83 & \textbf{0.15} & 0.61\\
& \textit{deepseek}  & \textbf{1.15} & \textbf{0.08} & \textbf{0.53} & \textbf{2.15} & \textbf{0.13} & \textbf{0.57} &\textbf{2.23} & \textbf{0.15} & \textbf{0.51}\\
\midrule
\multirow{4}{*}{\begin{sideways}\textit{w.o. meta}\end{sideways}} 
& \textit{4o-mini}   & 0.39 & 0.08 & 0.63 & 1.25 & 0.16 & \textbf{0.62} & 1.87 & 0.17 & 0.62\\
& \textit{o3-mini}   & \ul{1.41} & \ul{0.07} & \ul{0.59} &\ul{1.84} & \ul{0.16} & \ul{0.63} & \ul{2.01} & \textbf{0.15} & \textbf{0.53}\\
& \textit{gemini}    &  0.05 & 0.08 &  0.70 & 1.76 & 0.17 & 0.76 & 1.86 & \textbf{0.15} & 0.60\\
& \textit{deepseek}  &  \textbf{1.76} & \textbf{0.06} & \textbf{0.52}& \textbf{1.85} & \textbf{0.16} & \textbf{0.62} &\textbf{2.21} & \textbf{0.15} & \textbf{0.53}\\
\bottomrule
\end{tabular}
\caption{Evaluation on obfuscation. KL ($\uparrow$), SIM ($\downarrow$), and Verification Accuracy (ACC) ($\downarrow$) between the mimicked and original text.}
\label{tbl_obfuscation_mimicking}
\vspace{-5pt}
\end{table}

\noindent \textbf{Influence of Obfuscation.} To quantify AO, we employ a mimicking judge (\textit{4o-mini}) and a verification judge (\textit{o3-mini}). 
Table~\ref{tbl_obfuscation_mimicking} reports the KL and SIM between mimicked and original texts and the verification accuracy on original texts when using obfuscated texts as the ground truth. Overall, among all models, \textit{deepseek} consistently demonstrates the strongest obfuscation influence across all datasets, achieving the highest KL and lowest SIM scores. This indicates that its obfuscated outputs \textit{deviate the most from the original writing style}. 

In addition, the results also show that \textit{obfuscation without user metadata generally outperforms the versions that incorporate metadata}. This suggests that metadata may inadvertently constrain the models, making it more difficult to mask the original writing style. In other words, the mimicking judge can utilize the same user metadata to reconstruct the original author's writing style, making the obfuscation less impactful. Furthermore, the performance gap between the with-metadata and without-metadata settings is most pronounced in the Speech dataset, which features more well-known authors. This gap progressively narrows in the Quora and Essay datasets, reflecting a trend: \textit{it is easier to conceal the identity of less well-known authors, regardless of metadata inclusion}.

\renewcommand{\tabcolsep}{1.2pt}
\begin{table}[tb]
\footnotesize
\centering
\begin{tabular}{llcccccccccc}
\toprule
& \multirow{2}{*}{\textbf{Models}} 
& \multicolumn{3}{c}{\textbf{Speech}} 
& \multicolumn{3}{c}{\textbf{Quora}} 
& \multicolumn{3}{c}{\textbf{Essay}} \\
\cmidrule(lr){3-5} \cmidrule(lr){6-8} \cmidrule(lr){9-11}
& & KL & SIM & ACC
  & KL & SIM & ACC
  & KL  & SIM & ACC \\
\midrule
\multirow{4}{*}{\begin{sideways}\textit{w. meta}\end{sideways}} 
& \textit{4o-mini}   & \ul{3.25} & \textbf{0.05} & \ul{0.73} & \ul{2.51} & \ul{0.17} & 0.78 & \ul{2.32} & 0.20 & \ul{0.68} \\
& \textit{o3-mini}   & 2.95 & 0.06 & 0.70 & 2.30 & \ul{0.19} & 0.73 & 2.14 & 0.19 & 0.65 \\
& \textit{gemini}    & \textbf{3.29} & \textbf{0.05} & \textbf{0.87} & \textbf{3.20} & \textbf{0.15} & \textbf{0.89} & \textbf{2.98} & \textbf{0.18} & \textbf{0.71}\\
& \textit{deepseek}  & 2.95 & 0.07 & 0.65 & 2.18 & 0.18 & \ul{0.82} & 1.97 & 0.21 & 0.67\\
\midrule
\multirow{4}{*}{\begin{sideways}\textit{w.o. meta}\end{sideways}} 
& \textit{4o-mini}   & \textbf{3.32} & \ul{0.06} & \ul{0.70} & 2.13 & \ul{0.16} & 0.79 & \ul{2.16} & \ul{0.20} & \ul{0.63} \\
& \textit{o3-mini}   & 3.26 & \ul{0.06} & 0.62 & 2.24 & 0.19 & 0.64 & 1.98 & 0.22 & 0.60\\
& \textit{gemini}    &  \ul{3.28} & \textbf{0.05} & \textbf{0.82} & \textbf{2.48} & \textbf{0.15} & \textbf{0.87} & \textbf{2.79} &\textbf{0.19} & \textbf{0.69}\\
& \textit{deepseek}  & 2.58 & 0.07 & 0.59 & \ul{2.37} & 0.17 & \ul{0.81} & 2.03 & 0.22 & 0.62 \\

\bottomrule
\end{tabular}
\caption{Evaluation on mimicking. KL ($\uparrow$), SIM ($\downarrow$), and Verification Accuracy (ACC) ($\uparrow$) between the obfuscation and original text.}
\label{mimicking_evaluation}
\vspace{-5pt}
\end{table}

\vspace{3pt}
\noindent \textbf{Influence of Mimicking.} To quantify AM, we evaluate the mimicked texts using two distinct judges: an obfuscation judge (\textit{o3-mini}) and a verification judge (\textit{o3-mini}). Table~\ref{mimicking_evaluation} reveals several consistent trends across datasets. Gemini achieves the strongest overall performance in text obfuscation and verification, followed by 4o-mini, with Gemini leading in most KL ($\uparrow$), SIM ($\downarrow$), and ACC ($\uparrow$) metrics. \textit{Contrast with previous AO evaluation, incorporating user metadata to AM significantly enhances verification quality specially on Speech data}. Notably, the performance gap between settings with and without metadata narrows from well-known to lesser-known authors, suggesting that metadata plays a more critical role in capturing and disguising distinctive writing styles. Specifically, in the Speech dataset, the gap in KL divergence and SIM metrics between the metadata and without-metadata settings is substantially larger for AO than for AM. This implies that metadata is more influential in AO or that AO is generally more effective than AM. One possible explanation is that the input text contains many identifiable linguistic patterns, making it easier to alter (for obfuscation) than to replicate (for mimicking).

\vspace{3pt}
\noindent \textbf{Influence of Verification.} 
We construct noisy samples $\overline{C}_a$ by doing AV across the 4 models, which then serve as inputs for obfuscation and mimicking judge. Overall, \textit{o3-mini} achieves the highest precision and recall, with \textit{deepseek} showing strong recall, while \textit{4o-mini} and \textit{gemini} perform less effectively in AV.  We refer to Appendix.~\ref{precision_recall} for detailed setup and results.

\renewcommand{\tabcolsep}{2.5pt}
\begin{table}[tb]
\footnotesize
\centering
\begin{tabular}{llccccccccccc}
\toprule
& \multirow{3}{*}{\textbf{Models}} 
& \multicolumn{4}{c}{\textbf{$\overrightarrow{VO}$}} 
& \multicolumn{4}{c}{\textbf{$\overrightarrow{VM}$}} \\
\cmidrule(lr){3-6} \cmidrule(lr){7-10}
& & \multicolumn{2}{c}{\textbf{Speech}} & \multicolumn{2}{c}{\textbf{Quora}} &  \multicolumn{2}{c}{\textbf{Speech}} & \multicolumn{2}{c}{\textbf{Quora}} \\
\cmidrule(lr){3-4} \cmidrule(lr){5-6} \cmidrule(lr){7-8}
\cmidrule(lr){9-10}
& & KL & SIM & KL & SIM & KL & SIM 
  & KL & SIM \\
\midrule
\multirow{4}{*}{\begin{sideways}\textit{w. meta}\end{sideways}} 
& \textit{4o-mini}   & 1.47 & \ul{0.24} & 1.89 & 0.19  & 0.21 & \ul{0.33} & 0.39 & \ul{0.26}  \\
& \textit{o3-mini}   & \textbf{1.08} & \textbf{0.27} & \textbf{1.57} & \textbf{0.24} & \textbf{0.19} & \textbf{0.34} & \textbf{0.30} & \textbf{0.28} \\
& \textit{gemini}    & 1.65 & 0.22 & 1.80 & 0.18 & 0.22 & 0.30 & 0.40 & 0.25 \\
& \textit{deepseek}  & \ul{1.21} & \ul{0.24} & \ul{1.74} & \ul{0.21} & \ul{0.20} & \ul{0.33} & \ul{0.35} & \ul{0.26}  \\
\midrule
\multirow{4}{*}{\begin{sideways}\textit{w.o. meta}\end{sideways}} 
& \textit{4o-mini}   & 1.72 & \ul{0.22} & 1.91 & 0.17 & 0.34 & 0.29 & 0.41 & 0.25  \\
& \textit{o3-mini}   & \textbf{1.24} & \textbf{0.24} & \textbf{1.60} & \textbf{0.23} & \textbf{0.24} & \textbf{0.31} & \textbf{0.36} & \textbf{0.27}  \\
& \textit{gemini}    & 1.71 & 0.18 & 1.83 & 0.17 & 0.33 & 0.28 & 0.43 & 0.25 \\
& \textit{deepseek}  & \ul{1.45} & 0.21 & \ul{1.72} & \ul{0.20} & \ul{0.29} & \textbf{0.31} & \ul{0.38} & \ul{0.26} \\
\bottomrule
\end{tabular}
\caption{Evaluation on verification. KL ($\downarrow$) and SIM ($\uparrow$) measure similarity between two obfuscated texts. 
Full results are shown in Table \ref{full_verification_obfuscation_mimicking}.}
\label{verification_obfuscation_mimicking}
\vspace{-18pt}
\end{table}

Table \ref{verification_obfuscation_mimicking} reports how AV influences AO and AM when feeding AV with perfect (${C}_a$) and noisy samples ($\overline{C}_a$). Overall, models with higher precision, indicating fewer \textit{false positives} in $\overline{C}_a$ (Eq.~\ref{verification_o}, Eq.~\ref{verification_m}) and reduced noise in the few-shot ground truth, exhibiting smaller divergence between obfuscation texts generated with perfect and imperfect samples. This suggests that cleaner sample ground truth examples make the obfuscation texts more indistinguishable. Moreover, removing metadata during obfuscation amplifies the divergence between obfuscated texts, potentially because the obfuscation judge can utilize the metadata to force the obfuscated texts to be similar. Lastly, across datasets, the gap in KL and SIM becomes narrower as the author becomes less well-known, reflecting the \textit{diminishing influence of author-specific features in obfuscation}. 

In terms of $\overrightarrow{VM}$, overall, mimicked texts derived from ground-truth examples of LLMs with higher precision exhibit lower divergence, reflected by smaller KL and higher SIM, because higher precision reduces false positives and thus introduces less noise during the mimicking process. Additionally, \textit{AV's access to metadata consistently improves the AM judge’s ability to perform accurate text mimicking compared to settings without metadata, although this benefit diminishes as the authors become less well-known}. The reason might be LLMs' familiarity with famous people, and hence able to effectively utilize metadata.

\begin{figure}[tb]
    \centering
    \includegraphics[width=0.49\textwidth]{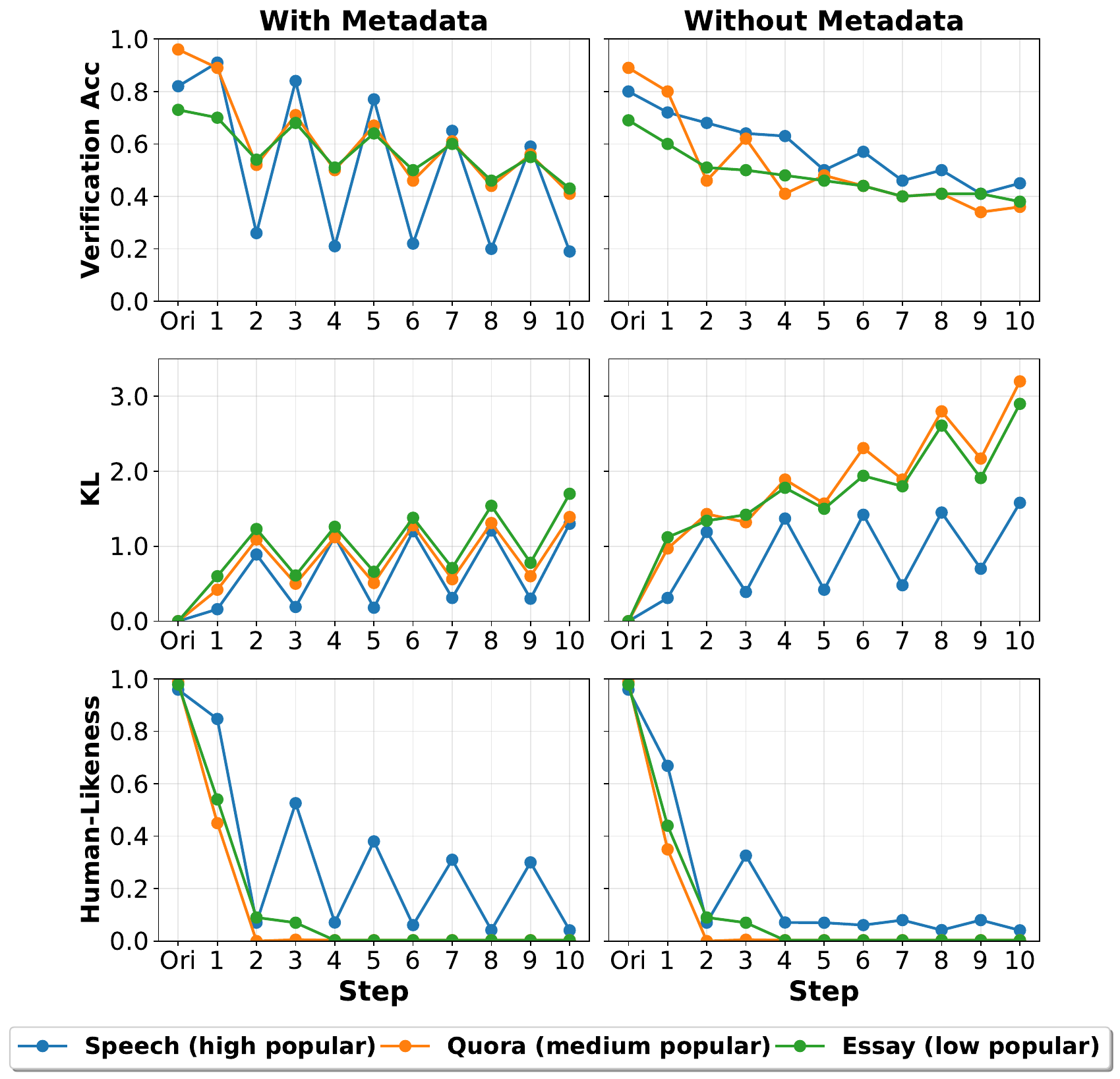}
    \caption{Verification accuracy ($\uparrow$), KL ($\downarrow$), and Human-likeness scores of mimicked and obfuscated texts compared to original texts across datasets, both with and without metadata. The x-axis represents the step order, ranging from 1 to 10 for 5 iterations \textit{alternating between AM$\rightarrow$AO$\rightarrow$AM$\rightarrow$...$\rightarrow$AO}. AV is used as an intermediate step after AO and does not generate any texts, so we hide it for clarity. We refer to Table~\ref{detail_loop_verification} for the detailed results.}
    \label{loop_evaluation}
    \vspace{-18pt}
\end{figure}

\subsection{Triplet-wise Interdependency (RQ2, RQ3)}
This section analyzes five \textit{iterative cycles} of AO, AM, and AV to evaluate how LLMs progressively shape stylometric patterns over time.
Without loss of generality, we begin with mimicking followed by obfuscation, as their outputs are iteratively used as inputs for the subsequent task throughout the evaluation process. 
An interesting observation is the emergence of zig-zag patterns in all plots in Fig.~\ref{loop_evaluation}, suggesting an ongoing ``tug-of-war'' between mimicking and obfuscation. Obfuscation appears to be more dominant, though the nature of this interplay varies depending on (1) the dataset and (2) the presence or absence of metadata.

\vspace{3pt}
\noindent \textbf{Authorship Verification.}
Overall, mimicking demonstrates the ability to recover the original text to some extent (first plot in Fig.~\ref{loop_evaluation}). However, its effectiveness diminishes over successive iterations, due to the cumulative noise introduced by repeated obfuscation steps, which makes it increasingly difficult for the mimicker to reconstruct the original content. This degradation is particularly evident in the Quora and Essay datasets, where mimicking accuracy drops sharply after the first iteration. In terms of obfuscation, we observe a substantial reduction in verification accuracy for the Speech dataset compared to Quora and Essay. This suggests that \textit{obfuscation is more effective when author identity is strongly encoded in the text}, as is the case for public figures whose speech styles are easily recognizable. Notably, \textit{removing metadata from AO/AM consistently decreases verification accuracy across all datasets and iterations}, further demonstrating the value of auxiliary information in authorship verification.

\noindent \textbf{Language Naturalness.}
Overall, KL divergence increases over iterations, mirroring verification trends and signaling growing linguistic drift from the original text (second plot in Fig.~\ref{loop_evaluation}). Mimicking degrades over time, especially without metadata, while obfuscation consistently drives text away from its original form. Mimicking works best on shorter, structured texts like Speech, whereas obfuscation excels on longer, more variable texts like Quora and Essay due to richer linguistic features for distortion.

\vspace{3pt}
\noindent \textbf{Anthropomorphism Analysis.}
We investigate whether generated text becomes more human-written or machine-generated through successive iterations of AM$\rightarrow$AO \cite{cheng2025dehumanizing}. To quantify this, we employ GPTZero\footnote{\url{https://gptzero.me/}}, one of the most popular commercial deepfake text detectors, to assess the degree to which a given text \textit{resembles human writing}. Fig.~\ref{loop_evaluation} reports the human-likeness score- GPTZero’s estimated probability that a given text is written by a human. The first mimicked texts often appear most human-like, especially on the Speech dataset, while obfuscated texts consistently score low. Mimicking after obfuscation can partially restore human-like style, but this effect \textit{fades over time as the text becomes increasingly machine-generated}. For Quora and Essay, texts generated after the second iteration are generally classified as machine-generated. This may be attributed to the lower popularity and variability in writing styles within these datasets, making it harder for mimicking models to recover stylistic patterns. 
Without metadata, this effect intensifies across all datasets, texts quickly adopt machine-like traits after two iterations, with minimal recovery by AM even in the Speech dataset.

\vspace{3pt}
\noindent \textbf{Topic Distribution.}
We analyze how mimicking and obfuscation alter topic distributions using LDA \cite{blei2003latent} and find that iterative authorship tasks gradually shift texts away from their original themes.
For instance, in the Speech dataset, the initial texts cover topics such as \textit{politics}, \textit{elections}, \textit{health/life}, \textit{war/terror}, and \textit{economy/jobs} are replaced by more generic, repetitive content over time. This degradation may result from the \textit{compounding effects of generation}, as LLMs tend to produce less specific and more repetitive content \cite{holtzman2019curious}. Detailed topic trends are in Appendix~\ref{detailed_lda_analysis}.

\section{Discussions}
\noindent \textbf{\textit{Relationship between Authors' Popularity and Metadata's Effectiveness.}}
Including metadata significantly boosts AV effectiveness, especially for well-known individuals, heightening privacy risks through easier re-identification or impersonation. Otherwise, lesser-known authors are less affected, indicating that popularity increases identifiability. While obfuscation helps, it does not reliably ensure anonymity. These results carry important implications for LLM providers like OpenAI, Google: (1) LLMs may unintentionally erode user privacy by leveraging publicly available or leaked metadata; second, (2) incorporating privacy-preserving mechanisms into authoring and editing tools; (3) providing transparency and safeguards around how metadata is used or inferred in LLM-driven authorship tasks.

\vspace{3pt}
\noindent \textbf{\textit{The Double-edged Sword of LLMs: Empowering Privacy or Enabling Threats?}}
LLMs are double-edged tools. On one hand, users can utilize LLMs for privacy-preserving purposes. For instance, whistleblowers or vulnerable individuals may rely on LLM-powered obfuscation tools to share sensitive content anonymously. On the other hand, the same technology can be misused for impersonation or misinformation. Our results show that LLMs can convincingly mimic writing styles, especially when metadata such as demographics is available, opening the door for social engineering attacks or deepfake text generation.
Therefore, individuals must be aware that their public user-generated content, even absent explicit identifiers, can leave behind implicit rich digital traces. This raises an urgent need for tools that proactively evaluate and adjust online writings to minimize their digital traces.

\vspace{3pt}
\noindent \textbf{\textit{Impersonation and Misuse at Scale.}}
The interplay between AO and AM reveals that obfuscated text can still be reverse-engineered by powerful LLMs, especially with demographic cues. This poses real risks: malicious actors could impersonate public figures or institutions at scale to spread misinformation. As a result, stronger authorship detection tools are essential to identify AI-generated impersonations and trace their origins.

\section{Conclusion}
In this work, we present a unified framework to evaluate how LLMs interact across authorship obfuscation, mimicking, and verification, highlighting task-specific strengths and the role of demographic metadata. Our analysis quantifies interdependencies among these tasks and shows that obfuscation generally dominates mimicking in disrupting authorial signals, though mimicking can partially recover stylistic traits over time. Notably, models with stronger reasoning excel at verification and style concealment but struggle to faithfully replicate an author’s unique voice.

\clearpage
\newpage
\section*{Acknowledgements}
We thank ChatGPT and Grammarly for their assistance in typo correction.

\section*{Limitation}
Despite presenting a comprehensive evaluation framework for the three core authorship privacy tasks—authorship verification, obfuscation, and mimicking—using diverse linguistic metrics across a range of real-world datasets, our study is limited by the absence of human-centered evaluation. While automated metrics offer scalability and consistency, incorporating human judgment would provide valuable insights into the perceived naturalness, fluency, and effectiveness of obfuscated or mimicked text. This is especially important in assessing whether generated text truly conceals authorship or convincingly imitates another writing style from a human perspective. Future work could benefit from human-in-the-loop studies to better align evaluation with real-world perceptions and practical usability.

\section*{Broader Impacts and Ethics Statement}
This work raises important ethical considerations in authorship privacy. While our framework helps evaluate and improve privacy-preserving techniques, it also reveals how LLMs can deanonymize writers or impersonate them, posing risks to vulnerable individuals and enabling potential misuse, such as spreading misinformation. We urge the development of safeguards, such as tools that warn users of identifiability risks and stronger detection systems for AI-generated content. All data used are publicly available and handled following ethical research standards.
\bibliography{main}

\clearpage
\newpage
\appendix

\setcounter{table}{0}
\renewcommand{\thetable}{A\arabic{table}}

\section{Appendix}
\label{sec:appendix}
\subsection{Practical applications on authorship privacy}
\label{practical_application}
In this section, we present some applications of our research questions related to real-world authorship privacy.

\paragraph{RQ1:} A practical application of this research question in authorship privacy is enabling users, such as whistleblowers, activists, or social media participants, to select the most suitable LLM for their goals. For instance, if a user seeks to mask their identity when writing sensitive content, the analysis can guide them toward models with strong authorship obfuscation (AO) performance. Conversely, a journalist or researcher aiming to emulate a public figure's writing style might benefit from models that excel in authorship mimicking (AM). Similarly, platforms concerned with detecting AI-generated or impersonated text can rely on models with high authorship verification (AV) accuracy. Thus, understanding isolated LLM performance informs the deployment of tailored models in real-world authorship privacy scenarios.

\paragraph{RQ2:} A practical application of this research question in authorship privacy lies in improving the design and security of multi-step text processing pipelines used in sensitive communications. 
Specifically, in scenarios like anonymous online forums, whistleblower disclosures, or secure messaging, texts often undergo multiple transformations—generation, obfuscation, and verification—each performed by different LLMs. 
Understanding how these models influence one another and the interdependencies that arise helps identify potential privacy risks, such as:
\begin{enumerate}[leftmargin=\dimexpr\parindent-0\labelwidth\relax,noitemsep,topsep=0pt]
    \item whether obfuscation techniques are truly effective in concealing an author's style. For instance, whistleblowers and journalists who rely on textual obfuscation to anonymize their writing may still be at risk if LLMs can reverse-engineer their original style, allowing adversaries to trace the obfuscated text back to them. 
    \item anonymizing sensitive documents, e.g, legal testimonies or medical records, where ensuring that downstream mimicking models cannot recover the original author’s style is critical for privacy protection. 
    \item evaluating the potential misuse of LLMs in impersonation attacks, such as forging stylistically similar content for deception or misinformation. 
    \item forensic investigations, where reliable verification must distinguish genuine statements from adversarially altered or mimicked texts. Additionally, content moderation systems can leverage these insights to detect and flag deceptive or impersonated content, enhancing online platform safety and trust.
\end{enumerate}

\paragraph{RQ3:} A practical application of this question is in developing robust authorship privacy tools that account for real-world scenarios where text undergoes multiple rounds of transformation. For instance, in environments like anonymous publishing platforms or secure communication channels, text might be repeatedly mimicked, obfuscated, and verified using different LLMs. Understanding how these iterative cycles influence each other helps identify how privacy can degrade or be preserved over successive edits. This knowledge allows designers to build more effective multi-stage pipelines that maintain author anonymity, prevent unintended leakage of writing style, and improve the reliability of verification methods, ultimately enhancing the security and trustworthiness of authorship privacy systems.

\subsection{Additional statistics on evaluation dataset}
\label{dataset_statistics}
We present the statistics on the evaluation dataset in Table 
 \ref{tab:statistic_evaluation_dataset} and \ref{tab:evaluation_dataset_attribute_dist}.

\renewcommand{\tabcolsep}{2pt}
\begin{table}[ht]
\centering
\small
\begin{tabular}{cccccc}
\toprule
\textbf{Dataset} & \textbf{\#} & \textbf{Avg} & \textbf{Avg doc.} & \textbf{Avg \#sen.} & \textbf{\#} \\
                 & \textbf{Exam} & \textbf{length} & \textbf{length} & \textbf{per doc.} & \textbf{Authors} \\
\midrule
\textbf{Speech}  & 5,172 & 58.20 & 17.44 & 3.34 & 3\\
\midrule
\textbf{Quora}  & 9,899  & 294.62 & 18.83 & 15.64 & 5\\
\midrule
\textbf{Essays}  & 154 & 225.87 & 9.43 & 7.24 & 3\\
\bottomrule
\end{tabular}
\caption{Statistics of the evaluation datasets.}
\label{tab:statistic_evaluation_dataset}
\end{table}

\begin{table}[ht]
\centering
\small
\begin{tabular}{lll}
\toprule
\textbf{Attribute} & \textbf{Value} & \textbf{Count} \\
\midrule
\multirow{5}{*}{CEFR}   
 & B1\_1 & 914 \\
 & B1\_2 & 881 \\
 & A2\_0 & 470 \\
 & B2\_0 & 231 \\
 & XX\_0 & 73 \\
\midrule
\multirow{4}{*}{Acad.\ Genre}  
 & Sciences \& Tech. & 1,034 \\
 & Social Sciences   & 762 \\
 & Humanities        & 674 \\
 & Life Sciences     & 99 \\
\midrule
\multirow{3}{*}{Lang.\ Env.}  
 & EFL & 1,886 \\
 & ESL & 610 \\
 & NS  & 73 \\
\midrule
\multirow{2}{*}{Sex}
 & F   & 1,430 \\
 & M   & 1,139 \\
\bottomrule
\end{tabular}
\caption{Distribution of author attributes across 2,569 learners.}
\label{tab:evaluation_dataset_attribute_dist}
\end{table}

\subsection{Prompt Construction}
\label{app_prompt_construction}
\paragraph{Author identification} can be generated based on the attributes of each learner, including sex, academic background, level of English proficiency, and country of origin, to build a more targeted background persona. For example: 
\textit{The author is female. Her academic background is in the Humanities. Her English proficiency level is CEFR B1 (lower). She is from Singapore, an ESL environment (English as a Second Language).} 
The prompt construction for mimicking, attribution, and obfuscation are written in Table \ref{prompt_construction}.

\renewcommand{\tabcolsep}{1pt}
\begin{table}[tb]
\footnotesize
\centering
\begin{tabular}{lcccccc}
\toprule
\multirow{2}{*}{\textbf{Models}} 
& \multicolumn{2}{c}{\textbf{Speech}} 
& \multicolumn{2}{c}{\textbf{Quora}} 
& \multicolumn{2}{c}{\textbf{Essay}} \\
\cmidrule(lr){2-3} \cmidrule(lr){4-5} \cmidrule(lr){6-7}
& {Precision} & {Recall} 
& {Precision} & {Recall} 
& {Precision} & {Recall} \\
\midrule

\textit{4o-mini}   & 0.36 & 0.50 & 0.36 & 0.50 & 0.33 & 0.50 \\
\textit{o3-mini}   & \textbf{0.67} & \textbf{0.80} & \textbf{0.54} & \textbf{0.70} & \textbf{0.50} & \textbf{0.70}\\
\textit{gemini}    & 0.36 & 0.50 & 0.33 & 0.50 & 0.27 & 0.40\\
\textit{deepseek}  & \ul{0.62} & \textbf{0.80} & \textbf{0.54} & \textbf{0.70} & \ul{0.43} &  \ul{0.60}\\
\bottomrule
\end{tabular}
\caption{Authorship verification precision and recall of the four LLMs on the Speech, Quora, and Essay datasets.}
\label{precision_recall_table}
\end{table}

\renewcommand{\tabcolsep}{1.5pt}
\setlist[itemize]{topsep=0pt, partopsep=0pt, leftmargin=-1pt}
\begin{table*}[t]
\setlength{\belowcaptionskip}{-0.1cm}
\footnotesize
\centering
\begin{tabular}{p{2cm} p{12cm}}
\toprule
     \textbf{Task}  & \textbf{Prompt} \\
     \midrule
     \multirow{12}{2cm}{\raggedright Verification} &
     \begin{itemize}
         \item System Prompt: You are a judge designed to verify the attribution of a human-author written text.

         \item Instruction: You are given sample texts including 5 writings from the author and 5 writings from others. Analyze the writing styles of the input text, disregarding the differences in topic and content. Reasoning based on linguistic features such as phrasal verbs, modal verbs, punctuation, rare words, affixes, quantities, humor, sarcasm, typographical errors, and misspellings. Your task is to verify if the input text was written by \{\textit{author name}\}. As output, exclusively return yes or no without any accompanying explanations or comments.
         
         \item Context: Here is some information about the author: \{\textit{author identification}\}. The 10 sample writings: \{\textit{sample text}\}.
         
         \item Task: The input text is: \{\textit{input text}\}.
     \end{itemize} \\
     \midrule
     \multirow{15}{2cm}{\raggedright Mimicking}  & \begin{itemize}
         \item System Prompt: You are an emulator designed to replicate the writing style of a human author.

         \item Instruction: You are given 5 sample writings from the author. The goal of this task is to mimic the author’s writing style while paying meticulous attention to lexical richness and diversity, sentence structure, punctuation style, special character style, expressions and idioms, overall tone, emotion, and mood, or any other relevant aspect of writing style established by the author. Your task is to generate a {\textit{\{avg\}}}-word continuation that seamlessly blends with the provided input text. Ensure that the continuation is indistinguishable from both the input text and the 5 sample writings by the author. As output, exclusively return the text completion without any accompanying explanations or comments.
         
         \item Context: Here is some information about the author: \{\textit{author identification}\}. The 5 sample writings from an author: \{\textit{sample text}\}.

         \item Task: The input text is: \{\textit{input text}\}.

     \end{itemize} \\
     \midrule
     \multirow{15}{2cm}{\raggedright Obfuscation} & 
     \begin{itemize}
        \item System Prompt: You are an emulator designed to hide the writing style of a human author. 
        
        \item Instruction: You are given 5 sample writings from an author. The goal of this task is to conceal the author's writing style by carefully modifying lexical richness and diversity, sentence structure, punctuation patterns, special character usage, expressions and idioms, overall tone, emotion, mood, and any other distinguishing stylistic elements. Your task is to generate {\textit{\{avg\}}}-word continuation that has writing style significantly different from the provided input text. Strive to make the rewritten text distinguishable from both the input text and the 5 sample writings by the author. As output, exclusively return the text completion without any accompanying explanations or comments.
        
         \item Context: Here is some information about the author: \{\textit{author identification}\}. The 5 sample writings from an author: \{\textit{sample text}\}.
         
         \item Task: The input text is: \{\textit{input text}\}.
         
     \end{itemize} \\
     \hline
\end{tabular}
\caption{Prompt construction for the 3 tasks to evaluate LLMs ability.} 
\label{prompt_construction}
\end{table*}

\subsection{Evaluation Metrics}
\label{evaluation_metrics}
\paragraph{Word Distribution.}  
We employ TF-IDF to quantify each word’s significance within a document relative to the entire corpus. TF counts word occurrences, while IDF down-weights common terms. We extract TF-IDF vectors from our text sources and compute cosine similarity to assess stylistic and thematic alignment.

\paragraph{Language Naturality.}  
Perplexity (PPL) evaluates how well a language model predicts a given text, with lower PPL reflecting greater confidence and closer adherence to learned linguistic patterns. Since LMs capture typical language structures from large corpora, PPL is a proxy for naturalness. Here, we fine-tune GPT-2 \cite{radford2019language} on the original corpus and compute text-level PPL for both human-written and generated texts.

\subsection{Detailed results on Precision and Recall}
\label{precision_recall}
We construct the imperfect ground truth examples $\overline{x}_p$ by sampling 20 examples from the original texts, including 10 from the author and 10 from others. The target model will be used to verify authorship. All the examples classified as correct verification will be used as the ground truth for the obfuscation and mimicking processes. Table \ref{precision_recall_table} shows detailed results on Precision and Recall.

\subsection{Additional results for VO and VM}
We present detailed evaluation results of VO and VM in Table \ref{full_verification_obfuscation_mimicking}.
\renewcommand{\tabcolsep}{2.5pt}
\begin{table*}[tb]
\footnotesize
\centering
\begin{tabular}{llccccccccccccccc}
\toprule
& \multirow{2}{*}{\textbf{Models}} 
& \multicolumn{6}{c}{\textbf{VO}} 
& \multicolumn{6}{c}{\textbf{VM}} \\
\cmidrule(lr){3-8} \cmidrule(lr){9-14}
& & \multicolumn{2}{c}{Speech} & \multicolumn{2}{c}{Quora} & \multicolumn{2}{c}{Essay}
  & \multicolumn{2}{c}{Speech} & \multicolumn{2}{c}{Quora} & \multicolumn{2}{c}{Essay} \\
\cmidrule(lr){3-4} \cmidrule(lr){5-6} \cmidrule(lr){7-8}
\cmidrule(lr){9-10} \cmidrule(lr){11-12} \cmidrule(lr){13-14}
& & KL & SIM & KL & SIM & KL & SIM 
  & KL & SIM & KL & SIM & KL & SIM \\
\midrule

\multirow{4}{*}{\begin{sideways}\textit{w. meta}\end{sideways}} 
& \textit{4o-mini}   & 1.47 & \ul{0.24} & 1.89 & 0.19 & 1.34 & 0.28 & 0.21 & \ul{0.33} & 0.39 & \ul{0.26} & 0.69 & \ul{0.18} \\
& \textit{o3-mini}   & \textbf{1.08} & \textbf{0.27} & \textbf{1.57} & \textbf{0.24} & \textbf{1.26} & \textbf{0.31} & \textbf{0.19} & \textbf{0.34} & \textbf{0.30} & \textbf{0.28} & \textbf{0.52} & \textbf{0.19} \\
& \textit{gemini}    & 1.65 & 0.22 & 1.80 & 0.18 & 1.51 & 0.28 & 0.22 & 0.30 & 0.40 & 0.25 & \ul{0.63} & 0.17 \\
& \textit{deepseek}  & \ul{1.21} & \ul{0.24} & \ul{1.74} & \ul{0.21} & \ul{1.32} & \ul{0.29} & \ul{0.20} & \ul{0.33} & \ul{0.35} & \ul{0.26} & 0.64 & 0.17 \\

\midrule

\multirow{4}{*}{\begin{sideways}\textit{wo. meta}\end{sideways}} 
& \textit{4o-mini}   & 1.72 & \ul{0.22} & 1.91 & 0.17 & 1.35 & 0.27 & 0.34 & 0.29 & 0.41 & 0.25 & 0.66 & 0.17 \\
& \textit{o3-mini}   & \textbf{1.24} & \textbf{0.24} & \textbf{1.60} & \textbf{0.23} & \textbf{1.32} & \textbf{0.29} & \textbf{0.24} & \textbf{0.31} & \textbf{0.36} & \textbf{0.27} & \textbf{0.54} & \textbf{0.18} \\
& \textit{gemini}    & 1.71 & 0.18 & 1.83 & 0.17 & 1.49 & \ul{0.28} & 0.33 & 0.28 & 0.43 & 0.25 & \ul{0.63} & \textbf{0.18} \\
& \textit{deepseek}  & \ul{1.45} & 0.21 & \ul{1.72} & \ul{0.20} & \ul{1.34} & \ul{0.28} & \ul{0.29} & \textbf{0.31} & \ul{0.38} & \ul{0.26} & 0.65 & 0.17 \\

\bottomrule
\end{tabular}
\caption{Merged results from both evaluations: \textbf{Verification Obfuscation} and \textbf{Verification Mimicking}. KL ($\downarrow$) and SIM ($\uparrow$) measure similarity between two obfuscated texts. \textbf{Bold} and \ul{underline} indicate best and second-best performance per category.}
\label{full_verification_obfuscation_mimicking}
\end{table*}

\subsection{Additional evaluation results on triplet-wise interdependency}
\label{additional_loop_eval}
We present detailed evaluation results on triplet-wise interdependency in Table \ref{detail_loop_verification}. 
\renewcommand{\tabcolsep}{1.5pt}
\begin{table*}[tb]
\footnotesize
\centering
\begin{tabular}{lccccccccccccccccccccccc}
\toprule
& & \multicolumn{11}{c}{\textbf{Verification}} 
  & \multicolumn{11}{c}{\textbf{KL}} \\
\cmidrule(lr){3-13} \cmidrule(lr){14-24}
& & \multirow{2}{*}{Original} 
& \multicolumn{2}{c}{{1}} 
& \multicolumn{2}{c}{{2}} 
& \multicolumn{2}{c}{{3}} 
& \multicolumn{2}{c}{{4}} 
& \multicolumn{2}{c}{{5}} 

& \multicolumn{2}{c}{{1}} 
& \multicolumn{2}{c}{{2}} 
& \multicolumn{2}{c}{{3}} 
& \multicolumn{2}{c}{{4}} 
& \multicolumn{2}{c}{{5}} \\
\cmidrule(lr){4-5} \cmidrule(lr){6-7} \cmidrule(lr){8-9} \cmidrule(lr){10-11} \cmidrule(lr){12-13}
\cmidrule(lr){14-15} \cmidrule(lr){16-17} \cmidrule(lr){18-19} \cmidrule(lr){20-21} \cmidrule(lr){22-23}
& & & AM & AO & AM & AO & AM & AO & AM & AO & AM & AO 
 & AM & AO & AM & AO & AM & AO & AM & AO & AM & AO \\
\midrule

\multirow{3}{*}{\begin{sideways}\textit{w meta}\end{sideways}} 
& \textit{Speech} & 0.82 & 0.91 & 0.26 & 0.84 & 0.21 & 0.77 & 0.22 & 0.65 & 0.20 & 0.59 & 0.19 & 0.16 & 0.89 & 0.19 & 1.13 & 0.18 & 1.20 & 0.31 & 1.21 & 0.30 & 1.30   \\
& \textit{Quora}  & 0.96 & 0.89 & 0.52 & 0.71 & 0.50 & 0.67 & 0.46 & 0.61 & 0.44 & 0.56 & 0.41 & 0.42 & 1.09 & 0.50 & 1.12 & 0.51 & 1.28 & 0.56 & 1.31 & 0.60 & 1.39 \\
& \textit{Essay}  & 0.73 & 0.60 & 0.54 & 0.58 & 0.51 & 0.50 & 0.49 & 0.46 & 0.45 & 0.46 & 0.43 & 0.60 & 1.23 & 0.61 & 1.26 & 0.66 & 1.38 & 0.71 & 1.54 & 0.78 & 1.70 \\
\midrule

\multirow{3}{*}{\begin{sideways}\textit{wo meta}\end{sideways}} 
& \textit{Speech} & 0.80 & 0.72 & 0.68 & 0.64 & 0.63 & 0.50 & 0.57 & 0.46 & 0.50 & 0.41 & 0.45 & 0.31 & 1.19 & 0.39 & 1.37 & 0.42 & 1.42 & 0.48 & 1.45 & 0.70 & 1.58  \\
& \textit{Quora} & 0.89 & 0.80 & 0.46 & 0.62 & 0.41 & 0.48 & 0.44 & 0.40 & 0.41 & 0.34 & 0.36 & 
0.97 & 1.43 & 1.32 & 1.89 & 1.57 & 2.31 & 1.89 & 2.80 & 2.17 & 3.20   \\
& \textit{Essay}  & 0.69 & 0.60 & 0.51 & 0.50 & 0.48 & 0.46 & 0.44 & 0.40 & 0.41 & 0.41 & 0.38 & 1.12 & 1.34 & 1.42 & 1.78 & 1.50 & 1.98 & 1.80 & 2.61 & 1.91 & 2.90 & \\
\bottomrule
\end{tabular}
\caption{Performance analysis across 5 iterations (AM: mimicking, AO: obfuscation) for Verification and KL Divergence metrics.}
\label{detail_loop_verification}
\end{table*}

\subsection{Detailed results on topic distribution analysis}
\label{detailed_lda_analysis}
From Table \ref{tab:lda_topics_original} to \ref{tab:lda_topics_round5_task2}, we show detailed results on topic distribution analysis on the mimicking and obfuscation process.
\renewcommand{\tabcolsep}{1.5pt}
\begin{table*}[tb]
\footnotesize
\centering
\begin{tabular}{cc}
\hline
\textbf{Topic} & \textbf{Top Words} \\
\hline
0 & day, election, going, people, help, votes, working, could, got, better \\
1 & weapons, tax, best, let, people, made, could, plan, give, think \\
2 & people, country, time, right, look, together, one, border, want, believe \\
3 & iraq, health, costs, people, team, looking, war, year, care, working \\
4 & people, jobs, american, time, america, states, think, right, work, put \\
5 & new, nation, america, american, years, right, peace, workers, great, drug \\
6 & want, people, terrorists, important, college, enforcement, asking, terror \\
7 & one, security, people, country, war, life, let, never, america, american \\
8 & going, government, economy, world, america, afghanistan, iraq, getting, history, go \\
9 & want, going, people, americans, think, true, test, save, health, support \\
\hline
\end{tabular}
\caption{Top 10 words for each LDA topic on the original Speech dataset}
\label{tab:lda_topics_original}
\end{table*}

\renewcommand{\tabcolsep}{1.5pt}
\begin{table*}[tb]
\footnotesize
\centering
\begin{tabular}{cc}
\hline
\textbf{Topic} & \textbf{Top Words} \\
\hline
0 & america, nation, people, great, believe, world, together, continue, better, means \\
1 & people, states, world, new, energy, afghan, united, best, take, working \\
2 & weapons, people, country, know, america, act, got, work, tough \\
3 & 1st, country, good, could, china, american, always, quality, going, people \\
4 & nation, iraq, america, united, security, safe, people, states, choose, issue \\
5 & people, going, new, americans, great, way, thank, american, want \\
6 & people, going, know, right, american, one, policy, get, great \\
7 & people, economy, going, world, american, country, families, great, challenges, nation \\
8 & want, good, working, people, continue, america, get, let, need \\
9 & america, going, know, day, country, people, give, future, nation \\
\hline
\end{tabular}
\caption{(Round1 Step1: Mimicking) Top 10 words for each LDA topic}
\label{tab:lda_topics_round1_task1_mimicking}
\end{table*}

\renewcommand{\tabcolsep}{1.5pt}
\begin{table*}[tb]
\footnotesize
\centering
\begin{tabular}{cc}
\hline
\textbf{Topic} & \textbf{Top Words} \\
\hline
0 & energy, progress, remains, across, people, iraq, together, let, ensure \\
1 & built, probability, plane, boeing, airbus, children, life \\
2 & ensuring, remain, nation, costs, moving, people, yet, accountability, financial \\
3 & one, ensuring, future, fostering, ensure, secure, american, communities, \\
4 & ensuring, essential, fostering, sustainable, future, progress, growth, efforts, economic \\
5 & one, world, unity, life, fostering, resilience, children, wage \\
6 & innovation, progress, challenges, yet, ensuring, fostering, remains, future, essential \\
7 & commitment, last, crucial, energy, year, world, legal, principles, stability \\
8 &  future, progress, let, together, challenges, forward, shared, innovation, resilience \\
9 & people, time, seemed, yet, dreams, distant, whispers, future, hope \\
\hline
\end{tabular}
\caption{(Round1 Step2: Obfuscation) Top 10 words for each LDA topic}
\label{tab:lda_topics_round1_task2_obfuscation}
\end{table*}

\renewcommand{\tabcolsep}{1.5pt}
\begin{table*}[tb]
\footnotesize
\centering
\begin{tabular}{cc}
\hline
\textbf{Topic} & \textbf{Top Words} \\
\hline
0 & people, america, time, one, great, care, future, american, could, talking \\
1 & world, going, future, let, great, build, america, job, look \\
2 & challenges, america, world, new, innovation, good, always, moment, americans, embracing \\
3 & people, america, let, country, opportunity, world, time, essential, american \\
4 & support, people, power, future, let, progress, collective, respect, together \\
5 & commitment, people, principles, trade, essential, future, nation, ensuring, dedication \\
6 & work, future, people, challenges, let, requires, together, vote, commitment \\
7 & america, together, need, nation, challenges, states, total, let, open \\
8 & people, going, country, great, america, let, bad, win, things \\
9 & nation, right, progress, values, back, believe, going, time, security \\
\hline
\end{tabular}
\caption{(Round2 Step1: Mimicking) Top 10 words for each LDA topic}
\label{tab:lda_topics_round2_task1_mimicking}
\end{table*}

\renewcommand{\tabcolsep}{1.5pt}
\begin{table*}[tb]
\footnotesize
\centering
\begin{tabular}{cc}
\hline
\textbf{Topic} & \textbf{Top Words} \\
\hline
0 & let, time, together, future, essential, ensure, yet, progress, resilience \\
1 & progress, forward, let, together, path, future, ensuring, yet, test \\
2 & progress, shared, challenges, future, together, unity, resilience, yet, innovation \\
3 & innovation, future, could, change, time, progress, challenges, resilience, ensuring \\
4 & future, let, progress, even, fostering, together, challenges, hope, path \\
5 & day, time, yet, relentless, lies, collective, life, fostering, decisions \\
6 & ’, offer, energy, greater, yet, dialogue, fostering, often, today, security \\
7 & people, fostering, progress, solutions, ensuring, future, innovation, challenges, efforts \\
8 &  progress, future, collective, resilience, let, together, fostering, solutions, efforts \\
9 & sustainable, fostering, growth, future, economic, innovation, ensuring, essential, together \\
\hline
\end{tabular}
\caption{(Round2 Step2: Obfuscation) Top 10 words for each LDA topic}
\label{tab:lda_topics_round2_task2}
\end{table*}

\renewcommand{\tabcolsep}{1.5pt}
\begin{table*}[tb]
\footnotesize
\centering
\begin{tabular}{cc}
\hline
\textbf{Topic} & \textbf{Top Words} \\
\hline
0 & jobs, future, american, people, look, time, let, lot, need \\
1 & future, nation, together, values, people, going, build, vital, need, american \\
2 & commitment, jobs, right, americans, unwavering, challenges, economy, fill, good, better \\
3 & world, economy, good, fight, prevail, got, forces, iraq \\
4 & together, security, great, strategy, even, america, need, economic, made \\
5 & people, future, work, america, nation, together, values, commitment, american \\
6 & get, progress, ahead, future, nation, true, americans, shared, tomorrow, day \\
7 & going, people, really, see, great, know, coming, thing, election \\
8 & challenges, together, face, world, nation, resolve, forward, future, let \\
9 & always, people, believe, right, nation, freedom, country, bad, working \\
\hline
\end{tabular}
\caption{(Round3 Step1: Mimicking) Top 10 words for each LDA topic}
\label{tab:lda_topics_round3_task1}
\end{table*}

\renewcommand{\tabcolsep}{1.5pt}
\begin{table*}[tb]
\footnotesize
\centering
\begin{tabular}{cc}
\hline
\textbf{Topic} & \textbf{Top Words} \\
\hline
0 & future, energy, collaboration, fostering, growth, demands, resilience, sustainable, economic \\
1 & ensuring, forward, innovation, together, progress, demands, stability, clear, economy \\
2 & together, progress, future, let, shared, collaboration, innovation, challenges, collective \\
3 & let, progress, yet, together, hope, future, resilience, world, time \\
4 & progress, path, let, future, yet, forward, together, demands, ahead \\
5 & future, together, hope, ensuring, efforts, values, unity, resilience, let \\
6 & quiet, world, shared, becomes, yet, month, america, key, let \\
7 & progress, future, challenges, forward, ensuring, innovation, resilience, let, time \\
8 & time, global, need, one, north, right, let, people \\
9 & future, fairness, ensuring, ensure, economic, together, everyone, fostering, collaboration \\\\
\hline
\end{tabular}
\caption{(Round3 Step2: Obfuscation) Top 10 words for each LDA topic}
\label{tab:lda_topics_round3_task2}
\end{table*}

\renewcommand{\tabcolsep}{1.5pt}
\begin{table*}[tb]
\footnotesize
\centering
\begin{tabular}{cc}
\hline
\textbf{Topic} & \textbf{Top Words} \\
\hline
0 & get, want, future, time, let, understand, ensuring, clear, situation \\
1 & world, people, let, american, america, challenges, values, stand, together \\
2 & people, right, help, innovation, prosperity, free, thing, economic, families \\
3 & going, future, people, need, change, country, long, better, day \\
4 & people, see, things, going, something, action, stand, disaster, let \\
5 & people, bad, great, ’, nation, america, time, see, going, country \\
6 & progress, journey, shaped, nation, spirit, something, always, human, going \\
7 & america, together, people, future, let, world, requires, true, get \\
8 & people, support, economy, know, work, great, world, time, done \\
9 & world, america, ahead, future, yet, let, resolve, hope, freedom \\
\hline
\end{tabular}
\caption{(Round4 Step1: Mimicking) Top 10 words for each LDA topic}
\label{tab:lda_topics_round4_task1}
\end{table*}

\renewcommand{\tabcolsep}{1.5pt}
\begin{table*}[tb]
\footnotesize
\centering
\begin{tabular}{cc}
\hline
\textbf{Topic} & \textbf{Top Words} \\
\hline
0 & guide, future, forward, commitment, essential, wage, next, protect \\
1 & path, time, something, forward, energy, address, economic, achieving \\
2 & let, solutions, forward, ?, progress, ahead, time, innovation, change \\
3 & yet, forward, time, small, step, care, health, forged, energy \\
4 & together, future, world, commitment, yet, challenges, let, across \\
5 & shared, peace, resilience, path, collaboration, forward, progress, fostering \\
6 & future, let, forward, progress, path, time, challenges, innovation, keep \\
7 & progress, together, yet, one, let, change, time, vision, path \\
8 & progress, together, resilience, challenges, essential, forward, future, collaboration, ensuring \\
9 & future, together, let, progress, everyone, fostering, ensuring, innovation, build \\
\hline
\end{tabular}
\caption{(Round4 Step2: Obfuscation) Top 10 words for each LDA topic}
\label{tab:lda_topics_round4_task2}
\end{table*}

\renewcommand{\tabcolsep}{1.5pt}
\begin{table*}[tb]
\footnotesize
\centering
\begin{tabular}{cc}
\hline
\textbf{Topic} & \textbf{Top Words} \\
\hline
0 & together, better, great, world, path, win, easy, opportunity, need \\
1 & let, america, future, country, time, get, believe \\
2 & get, jobs, let, american, america, need, means, investing \\
3 & american, world, let, unwavering, ahead, work, opportunity, people, everyone \\
4 & let, done, people, job, forward, time, get, citizens, keep \\
5 & people, america, let, get, know, country, great, right, american \\
6 & got, ta, room, forward, doubt, ahead, fight, open, freedom, stay \\
7 & people, american, work, always, america, ’, time, act, future, nation \\
8 & nation, let, best, got, stay, get, folks, sure \\
9 & going, people, know, work, really, together, believe, want, world \\
\hline
\end{tabular}
\caption{(Round5 Step1: Mimicking) Top 10 words for each LDA topic}
\label{tab:lda_topics_round5_task1}
\end{table*}

\renewcommand{\tabcolsep}{1.5pt}
\begin{table*}[tb]
\footnotesize
\centering
\begin{tabular}{cc}
\hline
\textbf{Topic} & \textbf{Top Words} \\
\hline
0 & progress, together, shared, resilience, future, challenges, let, fostering, ensuring \\
1 & lost, time, one, momentum, path, progress, america, became \\
2 & future, fostering, innovation, progress, together, embracing, let, resilience, ensuring \\
3 & future, progress, ensuring, let, together, challenges, innovation, collaboration, vision \\
4 & future, progress, together, yet, forward, resilience, remains, shared, collective \\
5 & let, progress, together, forward, step, commitment, action, ensure, future \\
6 & together, future, progress, let, ensuring, path, unity, shared, commitment \\
7 & ensuring, across, future, challenges, resilience, essential, without, forward, together \\
8 & fostering, let, remains, future, progress, approach, financial, essential, together \\
9 & progress, future, shared, forward, together, collaboration, challenges, yet, innovation \\
\hline
\end{tabular}
\caption{(Round5 Step2: Obfuscation) Top 10 words for each LDA topic}
\label{tab:lda_topics_round5_task2}
\end{table*}
\end{document}